\documentclass[lettersize,journal]{IEEEtran}
\usepackage{amsmath,amsfonts}
\usepackage{algorithm} 
\usepackage{algorithmicx}  
\usepackage{algpseudocode}

\usepackage{multirow}
\usepackage{array}
\usepackage[caption=false,font=normalsize,labelfont=sf,textfont=sf]{subfig}
\usepackage{textcomp}
\usepackage{stfloats}
\usepackage{url}
\usepackage{verbatim}
\usepackage{graphicx}
\usepackage{cite}
\usepackage{color}
\usepackage{makecell}
\usepackage[table]{xcolor}
\usepackage{bbding}
\usepackage{pifont}
\usepackage{booktabs}
\begin{document}

\title{Heterogeneous Graph Transformer for Multiple Tiny Object Tracking in RGB-T Videos\\ }
\author{ Qingyu Xu, Longguang Wang,  Weidong Sheng, Yingqian Wang, Chao Xiao, Chao Ma, Wei An
\thanks{This work was supported in part by the National Natural Science Foundation of China under Grant 62301601. }
\thanks{Qingyu Xu, Weidong Sheng, Yingqian Wang, Chao Xiao, Chao Ma and Wei An are with the College of Electronic Science and Technology, National University of Defense Technology (NUDT), China. Emails: \{xuqingyu, wangyingqian16, xiaochao12, machao0408, anwei\}@nudt.edu.cn, shengweidong1111@sohu.com.  (Corresponding authors: Longguang Wang; Weidong Sheng.)}
\thanks{Longguang Wang is with the Aviation University of Air Force, China. Email: {wanglongguang15@nudt.edu.cn})}}%

\markboth{Journal of \LaTeX\ Class Files,~Vol.~14, No.~8, August~2021}%
{Shell \MakeLowercase{\textit{et al.}}: A Sample Article Using IEEEtran.cls for IEEE Journals}


\maketitle

\begin{abstract}

Tracking multiple tiny objects is highly challenging due to their weak appearance and limited features. Existing multi-object tracking algorithms generally focus on single-modality scenes, and overlook the complementary characteristics of tiny objects captured by multiple remote sensors. To  enhance tracking performance by integrating complementary information from multiple sources, we propose a novel framework called {HGT-Track (Heterogeneous Graph Transformer based  Multi-Tiny-Object Tracking)}.
Specifically, we first employ a  Transformer-based encoder to embed images from different modalities. Subsequently, we utilize  Heterogeneous Graph Transformer  to aggregate spatial and temporal information from multiple modalities to generate detection and tracking features. Additionally, we introduce a target re-detection module (ReDet) to ensure tracklet continuity by maintaining consistency across different modalities.
Furthermore, this paper introduces the first benchmark VT-Tiny-MOT (Visible-Thermal Tiny Multi-Object Tracking) for RGB-T fused multiple tiny object tracking.
Extensive experiments are conducted on VT-Tiny-MOT, and the results have demonstrated the effectiveness of our method. Compared to other state-of-the-art methods, our method achieves better performance in terms of MOTA (Multiple-Object Tracking Accuracy) and ID-F1 score. The code and dataset will be made available at https://github.com/xuqingyu26/HGTMT.

\end{abstract}


\section{Introduction}

\IEEEPARstart{M}{ulti-object} tracking (MOT) plays a crucial role in various applications such as autonomous driving, airspace surveillance, and motion prediction \cite{luo2021exploring, dasika2022cb+, weng2021ptp}. With the advancement of UAV technology, drone-borne remote sensing systems have become an important tool for data collection, and served as a supplement to manned aircraft and satellite remote sensing systems. However, the detection results obtained from single-modality sensors are often unreliable in challenging situations such as low illumination, heavy occlusion, and haze \cite{michaelis2019benchmarking}. To address these issues, aggregating complementary information from multiple modalities has shown promising results. Among the various sensor combinations available for drones, visible and thermal cameras are popular choices that provide high-resolution visible and thermal videos for detecting and tracking tiny objects in the far distance. 

Recently, several single object tracking (SOT) methods \cite{li2016learning, wang2020cross, zhu2020quality, zhang2022visible} have been proposed for tracking targets in visible-thermal videos,  { and have made significant advancements. However, the tracking of multiple tiny objects in RGB-T videos remains inadequately explored due to the following challenges: 
\begin{enumerate}
	\item The tracking methods that can exploit complementary information from visible and thermal modalities are mainly SOT methods, which overlook the interaction between different targets. It is non-trivial to transfer the SOT method to MOT method.
	\item  The absence of paired RGB-T video datasets hinders the progress in MOT research. While there are datasets available for object detection and single object tracking in RGB-T videos, datasets with annotated target IDs suitable for multiple tiny object tracking are still lacking.
	\item  Maintaining tiny object trajectory is challenging due to the limited detection and data association accuracy.
\end{enumerate}}


\begin{figure*}[t]
	\centering
	\includegraphics[width=\textwidth]{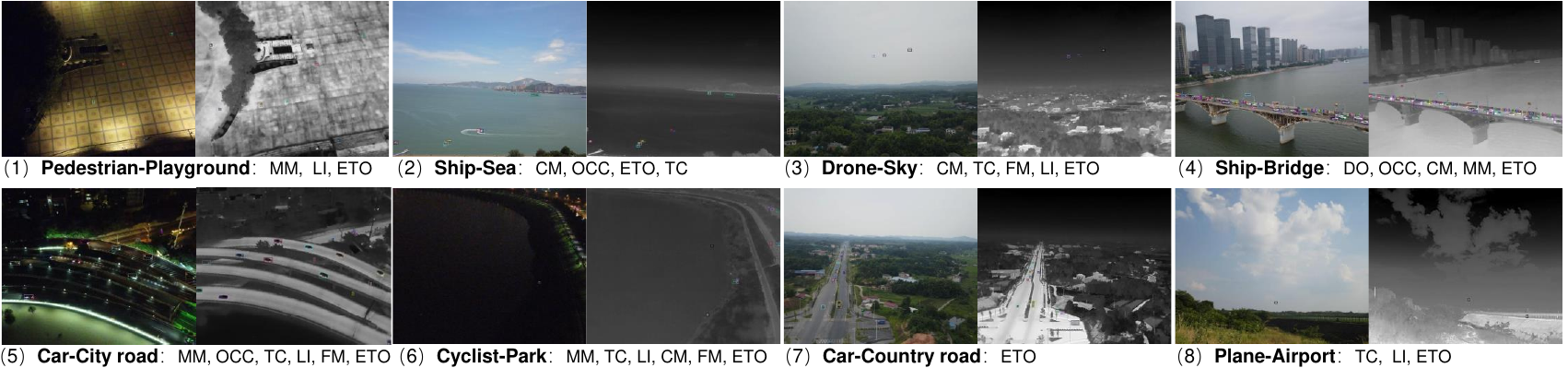}
	\caption{ {Examples of 7 kinds of targets captured under 8 main scenarios in the VT-Tiny-MOT dataset are provided, along with the annotated MOT challenges (the list of challenge attributes are reported in Table \ref{lay}). The involved challenges are MM (MisMatch), LI (Low Illumination), ETO (Extemely Tiny Object), CM (Camera Motion), OCC (OCClusion), TC (Thermal Crossover) and FM (Fast Move).}}
	\label{vtc}
\end{figure*}


 {Although significant progress has recently been made in MOT, the leading methods can not handle the RGB-T videos well. MOT methods typically follow the tracking-by-detection (TBD)  or the joint-detection-and-tracking (JDT) paradigm. The TBD methods \cite{wojke2017simple, bergmann2019tracking, xiao2021dsfnet, zhang2022bytetrack, cao2022observation} heavily rely on detection accuracy, making the tracking process susceptible to unstable detections in challenging scenarios. By using a bi-modal detector \cite{chen2022multimodal, sun2022drone},   the TBD methods can achieve a minor improvement due to the neglection of temporal information. Some JDT techniques \cite{kong2023cftracker, wang2021joint} have emerged to exploit the target temporal information.  However, these trackers' performance is limited in RGB-T videos as they are tailored to integrate homogeneous information and do not account for the domain gap between the visible and thermal modalities.}

 { 
	As illustrated in Fig. \ref{vtc}, the target state may differ in different modalities under diverse situations. For example, under low illumination (LI) scenarios, targets may be visible in the infrared spectrum and invisible in the visible spectrum, posing a challenge we refer to as `mismatch' denoted by `MM'.  Due to the discrepancies in target states and  sensor systems, simply merging data from two modalities could hinder tracking performance.}

 {Motivated by the success of Heterogeneous Graph Transformer in modeling  web-scale heterogeneous data, we seek to use Heterogeneous Graph Transformer \cite{hu2020heterogeneous} to  handle the  targets discrepancies in different modalities. 
}
In this paper, we present a novel approach called {HGT-Track (Heterogeneous Graph Transformer based  Multi-Tiny-Object Tracking)} that utilizes Heterogeneous Graph Transformer to construct the collaborative target representation. Unlike previous methods, {HGT-Track} models the relationship between different types of information by heterogeneous graph. Specifically, the targets at previous frame serve as tracking nodes, while the pixels at current frame serve as potential detection nodes, resulting in four types of nodes corresponding to  two modalities.  {HGT-Track} can efficiently integrate multi-modal information within the sparsely constructed heterogeneous graph. This integration allows {HGT-Track} to improve detection and tracking performance by leveraging the complementary information from the other modality. Notably, {HGT-Track} is specifically designed for multi-class small targets and is capable of recognizing objects in different scenes. Additionally, we propose a target re-detection module (ReDet) to ensure the trajectory continuity by maintaining consistency across different modalities.
 {In addition, we develop a new visible-thermal  dataset named VT-Tiny-MOT  for multiple tiny object tracking. The VT-Tiny-MOT dataset includes 115 paired sequences with 1.2 million manual tracking annotations. Noted that, targets are carefully annotated separately at the visible and thermal modalities.}

To summarize, our contributions  are as follows:
\begin{itemize}
	\item We propose an end-to-end Joint Detection and Tracking (JDT) framework, {HGT-Track}, for multiple tiny object tracking that effectively integrates visible and thermal information using Heterogeneous Graph Transformer. 
	\item We introduce a ReDet module, which improve the tracjectory continuity by maintaining consistency across different modalities.
	\item We construct a large-scale visible-thermal dataset named VT-Tiny-MOT for multiple tiny object tracking. Our proposed method {HGT-Track} achieves the best performance on thermal and visible videos of VT-Tiny-MOT as compared with other state-of-the-art trackers.
\end{itemize}

\section{Related Work} 
In this section, we briefly review three research streams that are closely related to our works.
\subsection{Multi-modal tracking}

With the increasing availability of various sensors, many multi-modal tracking methods \cite{li2019joint, roy2022multi, liu2023robust, nguyen2023multi} have been proposed to enhance the efficiency of tracking systems.  Lukezic et al. \cite{lukezic2019cdtb} investigate the utilization of depth information in visual object tracking.  However, depth sensors are limited to a confined range of 4-5 meters. To overcome this limitation, \cite{chen2022futr3d} employs LiDAR  and radar to capture accurate distance and angle information in autonomous driving scenarios. Compared with the sensors above, thermal sensors are more economically affordable and suitable for remote sensing applications. 

RGB-T fusion tracking can be categorized into pixel-level, feature-level, and decision-level fusion, depending on when the dual-modal information is fused. Li et al. \cite{li2016learning} propose an RGB-T fusion tracking method based on collaborative sparse representation at the decision level within a Bayesian filtering framework. To improve the efficiency of RGB-T object tracking, Zhai et al. \cite{zhai2019fast} introduce a low-rank constrained tracker to perform RGB-T tracking in the correlation filter framework. {Recently, several deep learning-based methods \cite{nam2016learning,  zhu2019dense, li2020challenge,  wang2022mfgnet, lu2022duality, long2019multi } have been proposed to enhance the information fusion in RGB-T tracking, which leverage the powerful capabilities of neural networks. MDNet \cite{nam2016learning} learns discriminative target representation with multiple domain-specific branches. Zhu et al. \cite{zhu2019dense} propose a method to reduce the noise and redundancy after the deep aggregation of multi-modality features. Li et al. \cite{long2019multi} propose an adaptive fusion module to improve information fusion efficiency. Though progress has been made, the deep-learning models still face problems such as large model sizes and high computation costs.} {Additionally, existing methods, whether traditional or deep learning-based,  are all focused on single object tracking without considering target number uncertainy.} Therefore, there is a necessity to explore efficient methods for multiple object tracking (MOT) in RGB-T videos.

\subsection{Multi-object tracking}
 {Current methods for Multiple Object Tracking (MOT) can be broadly categorized into two approaches: Tracking-by-Detection (TBD) and Joint Detection and Tracking (JDT). TBD treats tracking as a two-stage process involving detection followed by tracking. While modern detectors \cite{redmon2016you, zhou2019objects, zhu2020deformable} perform well in normal scenes, accurate tracking remains challenging in remote-sensed scenes due to weak target signals. To improve tracking performance in satellite videos, Xiao et al. \cite{xiao2021dsfnet} propose DSFNet, which generates accurate detections using 3D convolution to process video clips at the cost of processing speed. Some trackers \cite{zhang2022bytetrack, wang2022split, cao2022observation} optimize tracker and enhance data association accuracy by carefully designed procedures. In contrast, JDT methods \cite{sun2020transtrack, zhou2020tracking, zhang2021fairmot, chu2023transmot} merge the detector and tracker into a unified framework and perform joint optimization, leading to improved performance.}

 {To capture potential high-order correlations, graph-based methods \cite{papakis2020gcnnmatch, wang2021joint, weng2021ptp, he2022multi, chu2023transmot} have been proposed to exploit spatial and temporal information together. Wang et al. \cite{wang2021joint} propose a joint optimization method based on Graph Neural Networks (GNNs), which fuse spatial and temporal information to learn discriminative features. He et al. \cite{he2022multi} introduce a graph-based spatial-temporal reasoning procedure for online target tracking in satellite videos. Additionally, offline graph-based methods \cite{braso2020learning, wang2021track} have been proposed to improve tracking performance through global graph optimization. MPNTrack \cite{braso2020learning} learns a neural solver to handle data association in the graph domain, while LGMTracker \cite{wang2021track} tracks objects without appearance by encoding bounding boxes and tracklets through graph reasoning.  The methods above are tailored for multiple object tracking in single modality data without considering the discrepancy between different modalities.
Therefore, we extend the graph-based JDT framework for MOT in RGB-T videos by incorporating a modified Heterogeneous Graph Transformer module.}

\subsection{Small object tracking}
Small object detection and tracking are challenging \cite{marvasti2020comet, jiang2021anti, zhu2023tiny} due to low singal-to-noise ratio (SNR), particularly in military contexts, where remote objects often present as point or extended targets. Zhu et al. \cite{zhu2023tiny}  propose MKDNet, a novel method based on a multi-level knowledge distillation network, to enhance the performance of tiny object tracking. This suggests that tiny object tracking requires additional computing resources to strengthen target features.  Traditional tracking methods  \cite{ vo2013labeled, garcia2018poisson} address this issue within the framework of random finite sets (RFS). RFS methods often employ techniques such as hypothesis pruning or merging to alleviate the exponentially increased computation, which can lead to a trade-off in tracking accuracy.

Inspired by the capabilities of deep learning, there have been recent efforts to simultaneously detect and track small targets. For example, CenterTrack \cite{zhou2020tracking} detects and tracks target as point in a unified manner. In a different approach, He et al. \cite{he2022multi} utilize a graph-based spatiotemporal module to capture high-order relationships of small targets states between video frames in satellite videos. However, research on small target tracking using deep learning remains limited due to the lack of remote sensing data.

\section{Problem Formulation}\label{pf}
Multiple object tracking in RGB-T videos aims at integrating complementary information from both modalities to achieve robust tracking performance.

Assume that the state of object in the visible and thermal images is denoted by a set  ${X_k} = [{X}_k^v, {X}_k^t, q_k]$, where the superscript $v$ and $t$ represent the visible and thermal modality, respectively. The subscript  $k$ represents time.  Since the target may be neglected by detector, the target visibility represented by variable $q$ may have four possibilities: sensed by both, missed by both, or sensed by one certain modality. The state vector sets ${X}_k^v$  and $ {X}_k^t$ represent the objects state in corresponding modalities. The state vector of the $i^{th}$ object at time $k$ is defined as follows, 
\begin{equation}
        {X}_{k, i}^{v/t} = {[x \quad y \quad w \quad h ]^T},
\end{equation}
where $(x, y)$ represent the 2D position of the object in the camera image,  $(w, h)$ is the width and height of the target bounding box. 

The main task is to estimate $X_k$ with the sequence of image pairs ${\{ (I_i^v, I_i^t) \}}_{i=1}^{k}$,
 \begin{equation}
        (I_0^v, I_0^t), (I_1^v, I_1^t),..., (I_k^v, I_k^t) \Rightarrow {{X}_k},
\end{equation}
where the state set at time $k$ is estimated by processing the sequence of dual modality image pairs.%

For multi-modal JDT methods, the prior for detection is provided by both modalities:
 \begin{equation}
        {Prior} = {\rm Agg}(I_{k-1}^{v/t}, {\rm {X}}_{k-1}^{v/t}),
\end{equation}
where ${Prior}$ represent the aggregated prior information to enhance object detection at time $k$, and function $ {\rm Agg}(\cdot)$ is used to integrate multi-modal information.  Then, for each modality, the object can be detected with the prior information.
\begin{equation}
        Z^{v/t}_k = {\rm Det}(I_k^{v/t} \mid {Prior}),      
\end{equation}
where ${\rm Det}(\cdot)$ represents the detect function, $Z^v_k$ and $Z^t_k$ represent the detections at time $k$ in the visible and thermal modalities, respectively.

The tracking module takes the detection results from both modalities at time $k$ and previous target state at time $k-1$ as input, and outputs the estimation of target state vector at time $k$.
 \begin{equation}
        \tilde{X}^{v/t}_k = {\rm Track}({Z}_{k}^{v/t}, I_{k-1}^{v/t}, I_{k}^{v/t}),
\end{equation}
where function ${\rm Track(\cdot)}$ estimates the state vector  $\tilde{X}^v_k$ and $\tilde{X}^t_k$. The tracking module considers both temporal information and cross-modal information to improve the accuracy of state estimation. The tracking result can be used to aid in the association of detections with tracklets. Finally, the object state $X_k$  updates with the matched detections.

\section{Methodology}
\begin{figure*}[t]
\centering
\vspace{-.15in}
\includegraphics[width=0.9\textwidth]{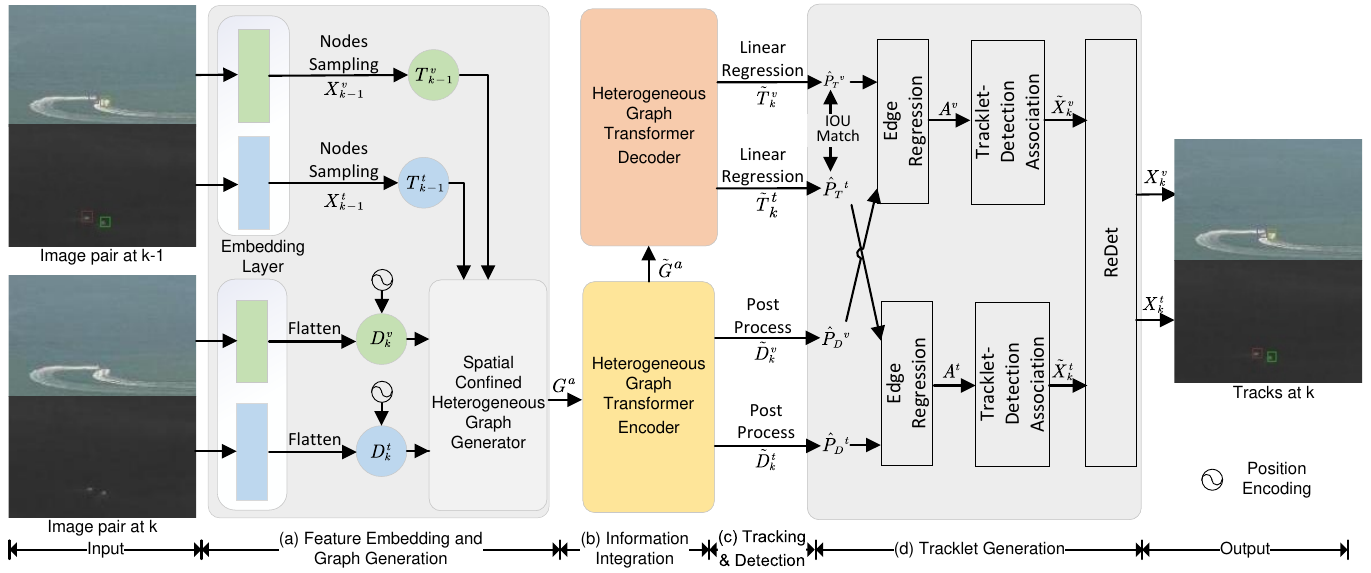}
\caption{The overall framework of Heterogeneous Graph Transformer based tracking method  {, which consists of four parts. (a) With the input of two paired visible and thermal images at time $k-1$ and $k$, we generate modal-specific feature through the embedding layer. Then, we build a heterogeneous graph $G^a$ considering the target difference between two modalities. (b) Next, we utilize the modlified HGT module for information integration, where the encoder output the detection feature and the decoder output the tracking feature. (c) A linear regression layer is used to generate the tracking offset, while the target detection is generated through conducting  a top-k post process on detection feature. (d) Finally, tracklet is generated and further refined through cross-modal detection matching and ReDet module. }}
\label{framework}
\end{figure*}

In this section, we present our {HGT-Track (Heterogeneous Graph Transformer based  Multi-Tiny-Object Tracking)}, which aims to effectively integrate spatial and temporal information from both modalities for joint online object detection and tracking. To achieve this, {HGT-Track} utilizes a Heterogeneous Graph Transformer (HGT) to aggregate target-specific information and enhance detection capabilities. Then, the generation of tracklets are conducted with the complementary feature.  An overview of the {HGT-Track} framework is illustrated in Fig. \ref{framework}.

We first introduce the Heterogeneous Graph Transformer-based encoder (Section \ref{II}). The HGT encoder fuses information by concurrently encoding four types of nodes and generating queries for detecting and tracking in both modalities. The dense detection queries are then used to regress and generate object detections. Subsequently, the Heterogeneous Graph Transformer decoder further decodes the sparse tracking queries to determine the tracking offset of targets in the current frame. In the tracklet generation part (Section \ref{tg}), we introduce a ReDet module to enhance the continuity of target trajectoy  aided by cross-modality object matching.
 \subsection{Feature Embedding and Graph Generation}

Given an RGB image $I^v_k \in \mathbb{R}^{W\times H \times 3}$  and a thermal image $I^t_k \in \mathbb{R}^{W\times H \times 1}$ obtained at time $k$, we initially project the images into modal-specific feature spaces using vanilla convolution.  Here,  $W$ and $H$  represent the width and height of the images, respectively. Next, we employ the parameter-shared Pyramid Vision Transformer (PVTv2) \cite{wang2021pyramid} to  {extract multiple scale features ${M_k^{v/t}} \in \mathbb{R}^{\frac{H}{r}\times\frac{W}{r}\times h}$ for each modality. In this context,  $r$ denotes the downscale ratio and $h$ denotes the feature dimension. Taking each pixel as potential detection, we reshape ${M_k^{v/t}}$ by the operation `Flatten' to  generate image dense detection queries ${D_k^{v/t}} \in \mathbb{R}^{\frac{HW}{r^2}\times h}$ for the two modalities.} 

 {
For the tracking feature, we treat the target center at previous frame as the target node. The tracking feature  ${T_{k-1}^{v/t}} \in \mathbb{R}^{{n^{v/t}_{k-1}}\times h}$ are sampled from  ${M_{k-1}^{v/t}} \in \mathbb{R}^{\frac{H}{r}\times\frac{W}{r}\times h}$, corresponding to the target nodes saved in ${X}_{k, i}^{v/t}$, namely `Nodes Sampling'.  Here,   $n^v_{k-1}$ and $n^t_{k-1}$ denote the  number of targets at time $k-1$ in corresponding modalities.}

The dense detection queries and sparse tracking queries serve as four types of graph nodes. Considering the bias between the two modalities, we use a heterogeneous graph  $G^a$ to model relationships among the various types of nodes. { We enable sparse interactions between the two modalities on nodes connected with $T_{k-1}^v$ and $T_{k-1}^t$. The spatial distance between all the linked nodes is smaller than a threshold $d$.
The graph $G^a$ is defined as follows:}
\begin{equation}
	{G}^a = \{ D_k, T_{k-1}, E_{DT}, E_{TT},  {E_{DH}} \},       
\end{equation}
{where  $D_k$  represents the set of potential detection nodes,  $T_{k-1}$ represents the set of tracking nodes, $E_{DT}$ represents the sets of temporal edges that link the tracking nodes with neighboring potential detection nodes of same modalities, $E_{TT}$ represents the sets of spatial edges that link the tracking nodes regardless of modalities, $ {E_{DH}}$ represents the sets of heterogeneous edges that link the tracking nodes with potential detection nodes of the other modality.
These three types of edges are defined for different purposes.  The temporal edges $E_{DT}$ are used to model the association relationship between tracking nodes and detection nodes. The weight of edges $E_{DT}$ will be further used to assign detections to target tracklets. $E_{TT}$ focuses on analyzing the differences between targets of different modalities, which will be utilized to match objects between the two modalities. $ {E_{DH}}$ establishes direct links between detection nodes and tracking nodes of different modalities, and is designed to incorporate temporal information from another modality to improve the performance of object detection.   }

 {We exclude the edge connecting detection nodes from different modalities due to two reasons. \textbf{First}, fully connecting detection nodes from different modalities will result in heavy computational costs. \textbf{Second}, pixel-wise fusion is unable to suppress the influence of bias between two modalities.}

\subsection{Information Integration}\label{II}
The defined $G^a$ with nodes and edges is then fed into the Heterogeneous Graph Transformer (HGT) encoder and decoder for information integration. 
\subsubsection{Heterogeneous Graph Transformer encoder}
\begin{figure}[tbp]
	\centering
	\includegraphics[width=7.8cm]{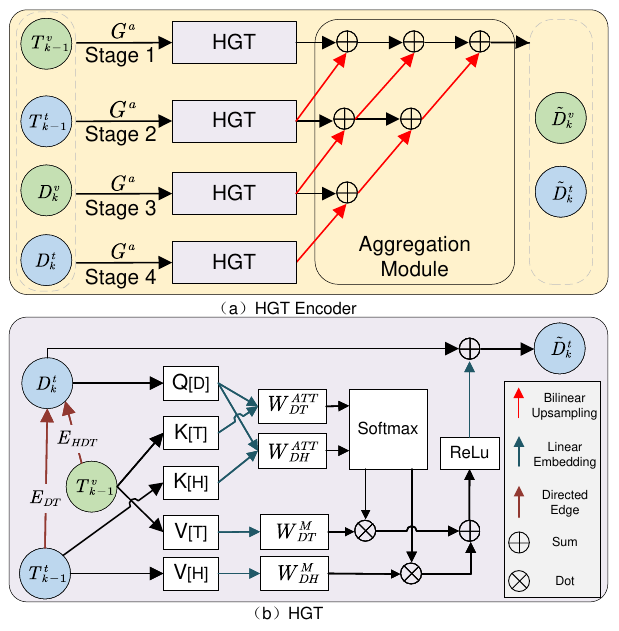}
	\caption{The structure of the Heterogeneous Graph Transformer encoder is illustrated in (a), and the details of the Heterogeneous Graph Transformer (HGT) are shown in (b). Multi-source information is aggregated to detection queries through HGT by setting the detection queries as target nodes and other types of nodes as source nodes. Then, the information is gradually integrated into $\tilde{D}$ from multiple stages of detection queries through the aggregation module.}
	\label{hgte}
\end{figure}

The Heterogeneous Graph Transformer  encoder, as illustrated in Fig. \ref{hgte} (a), is used to model the spatial and temporal correlations and learn cross-modal mapping by the modal-specific attention. Four stages of nodes $ D_k, T_{k-1}$ with the relationship defined by $G^a$ are fed into HGT separately. Since lower feature resolution means larger receptive fields \cite{li2022dense}, we strengthen the feature of region in $D_k$ with multi-scale features aggregated from other types of nodes linked by the directed edge $E_{DT}, E_{DH}$.     We illustrate the mechanism of Heterogeneous Graph Transformer in Fig. \ref{hgte} (b). 

 {Traditional Transformer \cite{vaswani2017attention} uses a single set projections for all kinds of nodes, while in our case each kind of node connection has a distinct set of projection weight following \cite{hu2020heterogeneous}.  Specifically, the distinct matrix $W_{\phi(e)}^{{\rm Att}}$ and $W_{\phi(e)}^{{\rm M}}$ are used to maintain the specific characteristics of different relations among the calculation of attention weights and message passing, where $\phi(e) \in \{ DT, DH, TT\}$ denotes the type of edge.  $W_{\phi(e)}^{{\rm Att}}$ and $W_{\phi(e)}^{{\rm M}}$ are optimized with the training of whole framework.}

 {Then, we calculate the heterogeneous mutual attention between different types of nodes. For example, consider nodes $D$ and $T$ from the same modality.} The nodes $T \in \mathbb{R}^d$ are projected through a linear layer to the $i^{th}$ head key vector $K^i \in \mathbb{R}^\frac{d}{h}$, and the parameters for different heads are not shared to maximize the distribution difference. The attention is calculated by:

\begin{equation}
	\begin{aligned}
		{\rm Att}(D, T) &= {\rm Softmax}(\mathop{\rm Concate}\limits_{i\in [1, h]}({\rm Att}^i({Q(D)}, {K(T)}))),   \\
		{\rm Att}^i(D, T) &= {\rm Linear}(K(T))W_{DT}^{{\rm Att}}{\rm Linear}(Q(D)),  
	\end{aligned}
\end{equation}   
where $h$ is the number of attention heads, and the attention matrix is calculated by concatenating the $h$ attention heads. The matrix $W_{DT}^{\text{ATT}} \in \mathbb{R}^{\frac{d}{h}\times\frac{d}{h}}$ is distinct to  { capture the semantic relations between the specific node type pairs.} The attention between $D$ and $T$ is generated by applying softmax to all neighboring nodes to ensure that the weights sum up to 1.

 {The message from node $T$ to $D$ incorporates the edge dependancy similarly by a distinct matrix $W_{DT}^{{\rm M}}$}:
 \begin{equation}
	\begin{aligned}
		&{\rm Message}(T) = {\rm Softmax}(\mathop{\rm Concate}\limits_{i\in [1, h]}{\rm Message}^i(K(T)))),   \\
		&{\rm Message}^i(K(T)) = {\rm Linear}(K(T))W_{DT}^{M},  
	\end{aligned}
\end{equation} 
 {where the i-th message head projects source node $K(T)$ to  the $i$-th message vector $K^i \in \mathbb{R}^\frac{d}{h}$. $K^i$ is then  multiplied with matrix $W_{DT}^{M} \in \mathbb{R}^{\frac{d}{h}\times\frac{d}{h}}$ to incorporate edge dependancy. The final message passing step is to concatenate all heads messages to get ${\rm Message}(T)$.}

The output of HGT is then generated by using the attention weight to aggregate messages from all types of source nodes. The updated $\tilde{D}$ is calculated as follows:
\begin{equation}
	\tilde{D} = \sum_{\forall T\in N(D), \forall e \in E(D, T)}({\rm Att}(D, T) \cdot {\rm Message}(T)),        
\end{equation}
where  $\sum$ aggregates the message from neighbor tracking nodes $N(D)$ with the attention weight.

The updated four stages of $\tilde{D}$ are further aggregated following the Iterative Data Aggregation (IDA) method \cite{yu2018deep}.  This aggregation process strengthens the detection query $D$ by incorporating complementary information from different modalities across frames.

\subsubsection{Heterogeneous Graph Transformer  decoder}

\begin{figure}[tbp]
	\centering
	\includegraphics[width=5cm]{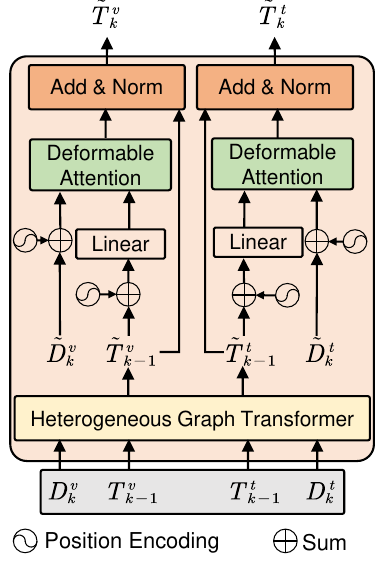}
	\caption{Illustration of the heterogeneous transformer decoder for the generation of tracking feature $\tilde{T}_k^v$ and $\tilde{T}_k^t$.  {Given the input of four types of nodes, namely  ${D_k^{v/t}}$ and ${T_{k-1}^{v/t}}$, we set ${T_{k-1}^{v/t}}$ as the target node. The Heterogeneous Transformer (HGT) integrates information from ${D_k^{v/t}}$ to generate  ${\tilde{T}_{k-1}^{v/t}}$. Subsequently, the tarcking feature ${\tilde{T}_{k}^{v/t}}$ are generated using deformable attention strengthening.}} 
	\label{hgmd}
\end{figure}

The HGT decoder, as shown in Fig. \ref{hgmd}, is utilized to decode the tracking feature, which is subsequently used to determine the displacement of the target from the current frame to time $k-1$. The HGT incorporates multi-source information into the tracking queries $\tilde{T}$ by setting them as target nodes.  The tracking queries $\tilde{T}$ and the current detection queries $\tilde{D}$ are encoded with position for better spatial interaction. Finally, tracking queries  $\tilde{T}_k^v$ and $\tilde{T}_k^t$   are learned separately through cross-deformable attention mechanism \cite{zhu2020deformable}.

\subsection{Object Detection and Tracking}
 {
For each modality, the decoded queries $\tilde{D}_k$ are post processed to generate detections. First, $\tilde{D}_k$ are reshaped back to $\mathbb{R}^{\frac{H}{r}\times\frac{W}{r}\times h}$, and then used as inputs to several branches consist of convolution layers to}   generate the object center heatmap $C_k \in \mathbb{R}^{\frac{W}{4}\times \frac{H}{4}\times 1}$, bounding box $S_k \in \mathbb{R}^{\frac{W}{4}\times \frac{H}{4}\times 2}$, and refine offset $R_k \in \mathbb{R}^{\frac{W}{4}\times \frac{H}{4}\times 2}$. Here, $S_k$ contains the height and width of the object's bounding box prediction map, which has a resolution that is $\frac{1}{4}$ of the original resolution. The target center $c_k$ are obtained by applying a threshold to $C_k$. The size of each object is extracted from $S_k$ corresponding to each position $c_k$. The set of detections is denoted as $Z_k=\{c_{k, i}, s_{k, i}\}_{i=1}^{N}$.

The tracking queries  $\tilde{T}$ are input into the tracking branch, which consists of two linear layers with ReLU activation, to regress the tracking offset $t_k \in \mathbb{R}^{\tilde{N}\times 2}$, where $\tilde{N}$ represents the number of targets at time $k-1$. Subsequently, the predicted positions of the objects $\tilde{c}_k = \{c_{k-1, i} + t_{k, i}\}_{i=1}^{\tilde{N}}$ can be computed using the predicted tracking offsets. These predicted positions are further used to assist in the detection assignment process.

\subsection{Tracklet Generation}\label{tg}
\subsubsection{Cross-modal detection matching}
Compared to normal single modality methods for MOT, {HGT-Track} performs cross-modal detection matching before detection-tracklet association in each modality. Given the detections $Z_k^t$ and $Z_k^v$, the distance $Dis$ between cross-modal detections is calculated using the Intersection over Union (IoU) metric:
\begin{equation}
	Dis(i, j) = 1 - {\rm IOU}(z_i^t, z_j^v),
	\label{a4}      
\end{equation}
where the function ${\rm IOU}$ computes the Intersection over Union between the $i^{th}$ detection  $z_{i}^t$ in $Z_k^t$  and the $j^{th}$ detection $z_{j}^v$ in $Z_k^v$. The detections are then matched using the Hungarian algorithm \cite{kuhn1955hungarian}. The variable $q_k$ of $X_k$ is updated based on the cross-modal matching result.
\subsubsection{Detection and tracklet association}
Then, we assign the detections to existing tracklets based on the edge affinity matrix $A$. The edge weights of $A$ are determined through regression using the edge feature. Specifically, the detection feature $U \in \mathbb{R}^{N \times h}$ is sampled   from  $\tilde{D}_k \in \mathbb{R}^{\frac{WH}{r^2}\times h}$ at the corresponding  detection positions. On the other hand, the tracking feature $V \in \mathbb{R}^{\tilde{N} \times h}$ corresponds to $\tilde{T}$ exactly. The edge feature $E$  is defined as:
\begin{equation}
	E_{ij} = U_i  - V_j,
	\label{eq14}
\end{equation}
where $U_i$ is the feature of the object at the $i^{th}$ detection node and $V_j$ is the feature of the $j^{th}$ tracking node. The affinity matrix $A$ of dimension $N \times \tilde{N} \times 1$ is then regressed from $E$.
 \begin{equation}
 	A = {\rm Sigmoid}({\rm Conv2}({\rm ReLU}({\rm Conv1}(E)))),
 	\label{fs}
 \end{equation}
where ${\rm Conv1}$ and ${\rm Conv2}$ represent two convolutional layers with a kernel size of 1, and  ${\rm Sigmoid}$ function is used to output the values of $A$ between 0 and 1. The matrix $A$ can be used to associate the detected objects with existing tracklets using the Hungarian algorithm \cite{kuhn1955hungarian}. If a newly detected object can be detected at 3 consecutive frames, it will initialize a new tracklet. Tracklets will be ended if they are lost for more than 20 frames.

\subsubsection{Re-detect target}
\begin{figure}[tbp]
	\centering
	\includegraphics[width=7cm]{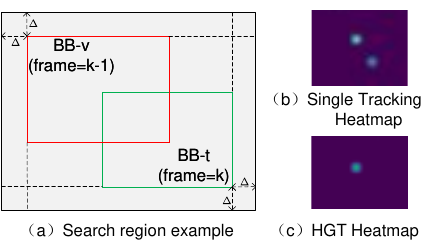}
	\caption{An example of the region area to re-detect the lost target. If the target is lost by the visible camera but can still be detected by the thermal camera, we initiate a re-detection procedure within a search region $SR$ colored by gray  at time $k$, as depicted in (a). This search region encompasses the union area of the target's bounding box at time $k-1$ in the visible camera and time $k$ in the thermal camera. In (b), we present the heatmap generated using ReDet to locate the lost target. Comparing it with the original HGT detection heatmap shown in (c), we can observe that even in a close-proximity scenario, another target can be accurately detected in the single tracking heatmap. }
	\label{tg1}
\end{figure}

The target is likely to have similar cross-modal tracklets if it can be detected by both modalities. In cases where no detection matches with  target  $X_{k-1, i}^v$, we attempt to re-detect the target  $X_{k, i}^v$ within the search region $SR$ of the visible frame, as illustrated in Fig. \ref{tg1}, but only if $X_i$ has been detected in both modalities. The search region is slightly larger than the union region of the bounding box of $X_{k-1, i}^v$ and $X_{k, i}^t$. Since target $X_{k, i}^v$ is too dim to be directly detected, we leverage the prior information from the other modality to provide complementary information. Specifically, we utilize the hybrid target feature output  from the HGT encoder to correlate with the search feature  extracted from $\tilde{D}_{k}^v$ corresponding to the search region, similar to single object tracking.  The pixels within the search region are treated as nodes representing potential detections, and the affinity matrix is computed following Eq. \ref{fs}, which serves as the correlation score map. The position with the highest score is further checked using our proposed two principles. The first principle checks whether the IOU distance between the  re-detected target $X_{k-1, i}^v$ and $X_{k, i}^v$ exceeds a certain threshold.  The second principle checks whether the detection score is above a filtered threshold  $\tilde{Det} $:
\begin{equation}
 	\tilde{Det} = \frac{Det}{1 + c_{k, i}^h},
\end{equation}
where $c_{k, i}^h$ represents the detection score of the lost target in the other modality, and ${Det} $ represents the  detection threshold of the detector. Consequently, the position of re-detected target is computed as:
\begin{equation}
	\begin{aligned}
		&(x, y) = \mathop{\arg\max}(A(E_{SR})), \\
		&if C_k(x, y) > \tilde{Det}, {\rm IOU}(Z_{k, i}, Z_{k-1, i}) > \tau,
	\end{aligned}
	\label{rd}
\end{equation}
where $E_{SR}$ represents the edge matrix between the lost target feature and the search feature, and $\tau$ is a selected threshold.
The details of the tracklet generation are summarized in Algorithm \ref{algorithm}.

%

\begin{algorithm}[t]
	\caption{The pipeline of Tracklet Generation}
	\label{algorithm}
	\begin{algorithmic}[1]
		\Require  Detections $Z_k$, object state $X_{k-1}$;
		\State Match the detections in both modalities  $Z^v_{k}$ and $Z^t_{k}$ depending on IOU distance;
		\For{each modality}
		\State Assign detections $Z_{k}$ to object $X_{k-1}$ using affinity matrix $A$.
		\State Update object state if it is matched with a detection. 
		\State Get the lost objects $LT$ if it is unmatched;
		\ForAll{$lt \in LT$}
		 \If{$X_{k-1, lt}$ is updated in one modality at step 4}
			 \State Re-detect $X_{k, lt}$ at the search region $SR$ using Eq.\ref{rd};
		 \EndIf
		 \State Update $X_{k, lt}$;
		\EndFor
		\EndFor
		\Ensure $X^v_{k}$ and $X^t_{k}$; 		  
	\end{algorithmic}
\end{algorithm}

\subsection{Loss Function}
The loss function consists of a detection loss term and a data matching loss term.

\subsubsection{Detection Loss}
 The detection loss is composed of $4$ parts including the center focal loss $L_{cf}$, the box size loss $L_{bs}$, the refinement loss $L_{r}$ and the tracking displacement loss $L_{td}$. The aforementioned four loss terms correspond to the classification branch and three regression branches, respectively. 

The center focal loss \cite{lin2017focal,law2018cornernet} $L_{cf}$ is used to train the heat map branch, which simultaneously penalize the classification error and localization error. The ground-truth heatmap $Y \in [0,1]^{\frac{H}{4}\times \frac{W}{4}\times 7}$, where $7$ is the number of object classes. 
 $L_{cf}$ is composed of the center focal loss $L_{cf}^t$ and $ L_{cf}^v$  in two modalities, and $ L_{cf}^v$ is computed as follows,
\begin{equation}
      \begin{aligned}
            &L_{cf}^v=\\
            & \frac{1}{N^v} \sum_{x y c}\left\{\begin{array}{ll}
                  \left(1-\hat{Y}_{x y c}\right)^\alpha \log \left(\hat{Y}_{x y c}\right) \qquad \qquad\text { if } Y_{x y c}=1 \\
                  \left(1-Y_{x y c}\right)^\beta\left(\hat{Y}_{x y c}\right)^\alpha \log \left(1-\hat{Y}_{x y c}\right)  \text { otherwise }
                  \end{array},\right.    \\
      \label{a8}     
\end{aligned} 
\end{equation}
Loss $L_{cf}^t$ is computed similarly to $ L_{cf}^v$.

For the remaining branches, losses are computed sparsely at the position of targets, i.e. the $i^{th}$ object at location $\mathbf{p}_i$ with the bounding box size $s_i$, the loss is computed as follows:
\begin{equation}
      \begin{aligned}
      L_{bs} &= \frac{1}{N^v}\sum_{i=1}^{N^v}|\hat{S}^v_{\mathbf{p}_i} - {s}_i^v| + \frac{1}{N^t}\sum_{i=1}^{N^t}|\hat{S}^t_{\mathbf{p}_i} - {s}_i^t|,
      \label{a10}     
\end{aligned} 
\end{equation}
Analogously to $L_{bs}$, the refinement loss $L_{r}$ and the tracking displacement loss $L_{td}$ are also computed at the center position of targets. The overall detection loss is the weighted sum of the four loss items. 

\subsubsection{Data matching loss}
In the detection and tracklet association part, the affinity matrix $A$  is regressed from the edge feature $E$. In the training stage,  the target tracking nodes $\{\tilde{T}_{k, i}\}_{i=1}^N$ and  the $N$  detection nodes are sampled from $\tilde{D}$ with the groundtruth positions.  The edge feature $E$ as Equation \ref{eq14} for the regression of $A$. 

The data matching loss $L_{match}$ is used to encourage matrix $A$ to approach an identity matrix $I \in \mathbb{R}^{N\times N}$. $L_{match}$ is formulated as the combination of the binary cross entropy loss $L_{BCE}$ and the cross entropy loss $L_{CE}$.

The loss  $L_{BCE}$ is computed pixel-wisely on $A$ to supervise the binary classification problem of the edges. It encourages $A$ to be close to either 0 or 1.   The loss $L_{CE}$ is computed to encourage each row and column of $A$ to be a one-hot vector.

\begin{equation}
      \begin{aligned}
     L_{match} = &\sum_{i\in\{t, v\}}L_{BCE}^i+L_{CE}^i,\\
      L_{BCE} = & -\frac{1}{N^2}\sum_{i=1}^{N}\sum_{j=1}^{N}I_{ij}\log (A_{ij}) + (1-A_{ij})\log (1-A_{ij}),\\
      L_{CE} = &-\frac{1}{N}\sum_{i=1}^{N}A_{ij}\log(\frac{{\rm e}^{A_{ij}}}{\sum_{i=1}^{N} {\rm e}^{A_{ij}}})\\
      & -\frac{1}{N} \sum_{j=1}^{N}A_{ij}\log(\frac{{\rm e}^{A_{ij}}}{\sum_{j=1}^{N} {\rm e}^{A_{ij}}}),
      \label{a9}     
\end{aligned} 
\end{equation}
where $N$ is the number of targets that simultaneously appear in one modality. 
 $L_{match}$ represents the loss values from two modalities. $L_{CE}$ is computed per row and column.

\section{Visible-Thermal Tiny Multiple Object Tracking Benchmark}
In this section, we introduce the newly collected visible-thermal tiny multiple object tracking  (VT-Tiny-MOT) benchmark, including  dataset construction with statistics analysis and baseline methods with both single-modality and dual-modality inputs.
\subsection{Data Collection and Annotations}
 The visible and thermal videos were simultaneously recorded  using visible and thermal cameras mounted on a professional UAV (DJI Mavic 2). Each video pair consists of a visible video and a thermal video. To ensure accurate alignment between the two modalities, a homography matrix is computed using  Zhang's method \cite{zhang2000flexible} and applied to align the remaining frames of each video pair. However, due to two main factors, the registration accuracy is limited  {, and there exists bias between each video pairs}.
 
 {\textbf{First}}, the vertical arrangement of the cameras on the UAV leads to variations in the depth of field (Dof), which violates the assumptions made in \cite{zhang2000flexible}.  {\textbf{Second}, the Zhang's method \cite{zhang2000flexible} is  ineffective in handling the disparity variations in stereo image pairs \cite{wang2020parallax}, especially in cases of texture-lacking images.}  Despite these limitations, the dataset has been refined to achieve a frame rate of $15$ FPS and a resolution of  $512\times 640$ pixels.
 

To capture diverse scenarios, the video pairs in our dataset were recorded in eight distinct outdoor scenes: sea, lake, bridge, city road, country road, playground, airport, and sky, under different seasons, weathers, and illuminations. The recording UAV was operated in three different states: cruise, hover, and ground. Each state determined the height and motion style of the camera. Further details regarding the challenges posed by these scenarios will be discussed in Section \ref{statis}.

Ground truth annotations for objects in the visible and thermal modalities are provided separately in our dataset. The annotations include the location, category, and tracking ID of each object, following the label format specified by the MOT challenge \cite{dendorfer2020mot20}. The visible and thermal annotations are paired with a one-to-one correspondence, except for extreme conditions such as low illumination. The tracking IDs are consistent across different modalities,  enabling comprehensive evaluation of tracking methods. To ensure the quality of the data annotations, professional annotators were employed, and a thorough verification process was conducted. The annotations undergo frame-by-frame evaluation by a minimum of three assessors to minimize errors and inconsistencies.


\subsection{Statistics}\label{statis}
\begin{figure}[tbp]
      \centering
      \includegraphics[width=\linewidth]{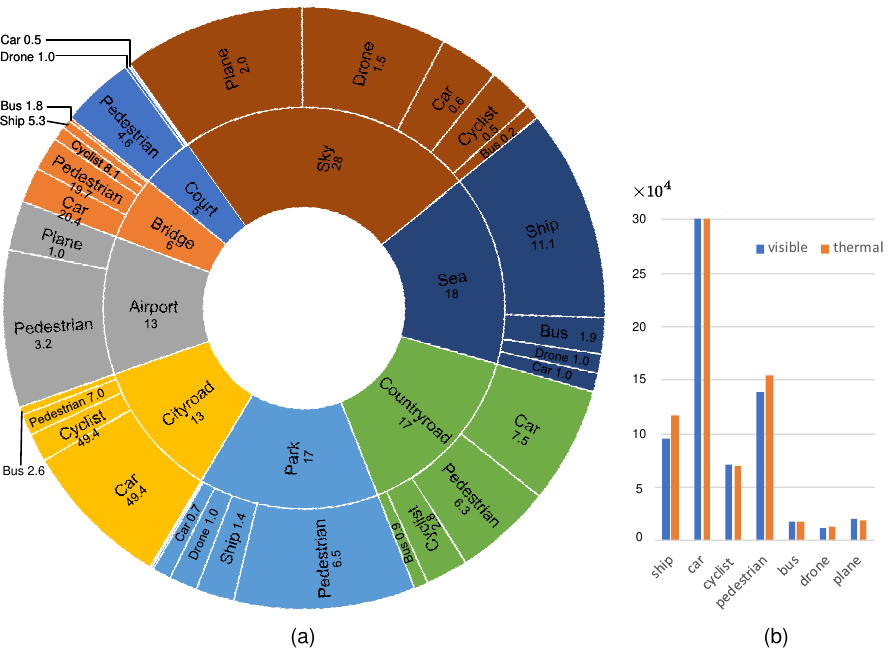}
      \caption{Statistics of the VT-Tiny-MOT dataset.  {(a) Distribution of scene (inner) with sequence numbers annotated and object category (outer) with average target number per frame annotated. (b) The target distribution in each modality.}}
      \label{ds}
      \end{figure}

The VT-Tiny-MOT dataset comprises a total of 115 video pairs, capturing a diverse range of circumstances, as illustrated in Fig. \ref{vtc}. The dataset contains 5208 instances of objects across seven different categories: ship, pedestrian, cyclist, car, bus, drone, and plane.  {The scene distribution and relevant target distribution with average numbers per frame are illustrated in Fig. \ref{ds} (a). The target type is highly correlated with the scene category (ship accounts for 97\% in `Sea' scene and car accounts for 73\% in `Cityroad' scene).  Due to variations in target states across modalities, targets were annotated separately for each modality, and the target distribution for each modality is depicted in Fig. \ref{ds} (b).}

To provide a context for the dataset, we compare its statistics with other RGB-T (visible-thermal) and small object tracking (SOT) datasets, as presented in Table \ref{stb}. This comparison allows researchers to understand the uniqueness and characteristics of the VT-Tiny-MOT dataset as compared to existing datasets.

Additionally, we identify ten specific challenges in the dataset, which are detailed in Table \ref{lay}. These challenges encompass various factors such as occlusion, low illumination, camera motion and others, which pose difficulties for object tracking algorithms.  We analyze the dataset from the following aspects.

\subsubsection{Small object} The presence of small targets poses a significant challenge due to their limited size and lack of rich shape and texture information. Compared to normal-sized objects, small targets are more difficult to perceive and track accurately. Specifically, the objects in the VT-Tiny-MOT dataset are mainly small targets, following the definition in \cite{lin2014microsoft}.  {As show in Table \ref{lay}, there are over $95.83\%$  of the targets being smaller than 1024 pixels. }
\subsubsection{Occlusion}
Occlusion occurs when targets are partially or fully obscured by the environment or other targets. For example, targets on city roads may be occluded by trees or buildings. Additionally, group targets such as helicopters can occlude each other. It is important to note that occlusion can also occur asymmetrically between the two modalities. In other words, a target may be occluded in one modality while still visible in the other modality. This introduces additional challenges for tracking algorithms, as they need to handle occlusion scenarios and effectively utilize the available modalities to maintain accurate object tracking.
\subsubsection{Modality mismatch} Due to the possibility of target loss in one modality, the annotations of a target in the two modalities do not always have an exact one-to-one correspondence. Modality mismatch, as a challenge, is closely related to two specific challenges, Low Illumination (LI) and Thermal Crossover.  In scenarios with low illumination, targets are difficult to detect by the visible camera, but the thermal information can aid in perceiving the targets. When an object has a similar temperature to the background, integrating the object information from the visible camera becomes useful in discriminating the object from the background. Despite the modality mismatch challenge, the overall quantities of visible and thermal annotations are comparable.

\begin{table*}

\tabcolsep=0.016\linewidth
\renewcommand\arraystretch{1.2}
\caption{Statistic comparison among existing RGB-T and Small Object Tracking (SOT) datasets. ``Cat" represents the number of target categories. ``Small Target" and ``Multi Targets" represent whether the captured targets of the dataset are small and multiple, respectively.}
\centering
\label{stb}
\begin{tabular}{c|c|c|c|c|c|c|c|c|c|c}
\hline
 \multirow{2}*{Benchmark}& \multirow{2}*{Booketitle}& \multirow{2}*{Year}& {Platform}&\multirow{2}*{Seq.}&{Total}&\multirow{2}*{Anno.}&\multirow{2}*{Resolution}&{Target} & \multirow{2}*{Small}& \multirow{2}*{Multiple}\\
 						&	&					&{Frames}	 &					&{Frames}&&&{Caotegories}&&\\
\hline
GTOT \cite{li2016learning}	&TIP&2016				&Ground	&50     &7.8K &7.8K	&$384\times 288$		&4	&$\times$		&$\times$ \\
VOT-RGBT \cite{kristan2019seventh}	&ICCVW&2019			&Ground	&60     &40K &40K	&$630\times 460$		&13	&$\times$		&$\times$ \\
RGBT234 \cite{li2019rgb}			&PR&2019		&Ground	&234    &233.8K&233.8K	&$630\times 460$		&22	&$\times$		&$\times$ \\
LasHeR \cite{li2021lasher}	&TIP&2021				&Ground	&1224   &6.7M&6.7M	&$630\times 480$		&32	&$\times$		&$\times$ \\
Anti-UAV \cite{jiang2021anti}		&TPAMI&2021		        &Ground	&318    &7.8K&7.8K	&$640\times 512$		&1	&$\checkmark$	&$\times$ \\
VTUAV  \cite{zhang2022visible}		&CVPR&2022			&UAV    &500	&3.3M&326K	&$1920\times 1080$	&13	&$\checkmark$	&$\times$ \\
\rowcolor{gray!20}VT-Tiny-MOT(ours)	  &-&-                              &UAV    &115    &93K&1.2M	&$512 \times 640$ 	&7	&$\checkmark$	&$\checkmark$\\

\hline

\end{tabular}
\end{table*}

\begin{table*}
	
	\renewcommand\arraystretch{1.2}
	\caption{List of the Attributes Annotated to the dataset VT-Tiny-MOT.  {We also  report the numbers of the sequences and annotations in VT-Tiny-MOT concerning the corresponding challenge at the `Seq' column and the `Anns' column, respectively.}}
	\label{lay}
	\centering
	\scriptsize
	\begin{tabular}{|c|c|c|c|c|p{9cm}|}
		\hline
		
		Modality& Attribute & Full Name  & {Seq}& {Anns} &Description \\
		\hline
		\hline
		\multirow{9}*{RGB}& OCC&Occlusion& {36} & { 0.1M }&The target is partially or fully occluded.\\
		&ETO&Extremely Tiny Object& {105}&  {0.45M} &The number of pixels in the ground truth bounding box is less than 64.\\
		&TO &Tiny Object& {110} &  {0.51M} &The number of pixels in the ground truth bounding box is between 64 and 256.\\
		&SO&Small Object & {71} & {0.19M}  &The number of pixels in the ground truth bounding box is between 256 and 1024.\\
		&DO&Dense Object& {10} &  {0.32M} &Targets densely appear in a region.\\
		&LI&Low Illumination& {55} &  {0.16M} &The illumination in the target region is low.\\
		&FM&Fast Move& {63} &  {0.01M} &Targets move beyond 10 pixels between two consecutive frames.\\
		&MM&Mismatch& {92} & {0.13M}  &Targets do not appear simultaneously in both modalities.\\
		&CM&Camera Motion& {17} & {365} &Camera moves when capturing the video.\\
		
		\hline
		
		Thermal&TC&Thermal Crossover& {55} &  {0.16M} &The target has similar temperature with other objects or background.\\
		\hline
	\end{tabular}
\end{table*}

\subsection{ {Baseline Methods}}
 {For evaluating the proposed method {HGT-Track} and providing a comprehensive evaluation benchmark, we include some popular tracking methods as baselines into our VT-Tiny-MOT benchmark.}

 {On one hand, there are 11 state-of-the-art single-modality trackers presented to demonstrate their performance in challenging scenarios, including  DeepSORT \cite{wojke2017simple}, Tracktor \cite{bergmann2019tracking}, ByteTrack \cite{zhang2022bytetrack}, OCSORT \cite{cao2022observation}, DSFNet \cite{xiao2021dsfnet}, MPNTrack \cite{braso2020learning}, CenterTrack \cite{zhou2020tracking}, FairMOT \cite{zhang2021fairmot}, TransCenter \cite{xu2022transcenter}, TraDes \cite{wu2021track} and GSDT \cite{wang2021joint}. On the other hand, we also implement 3 visible-thermal trackers by integrating bi-modal detectors ProbEn \cite{chen2022multimodal} and UA-CMDet \cite{sun2022drone} within DeepSORT, referred to as `ProbEn-E+SORT', `ProbEn-M+SORT', and `UA-CMDet+SORT', where `ProbEn-E' indicates the early fusion version of ProbEn, and `ProbEn-M' denotes the middle fusion version. Note that,   Faster R-CNN \cite{girshick2015fast} is utilized as the detector in TBD methods \cite{wojke2017simple, bergmann2019tracking, zhang2022bytetrack, cao2022observation, braso2020learning}. For the JDT methods \cite{zhou2020tracking, zhang2021fairmot, xu2022transcenter, wu2021track, wang2021joint}, training and testing are conducted following the protocols outlined in the respective original papers. The methods above are trained with various modalities, which is indicated in column `Mod'.}
\section{Experiments}
In this section, we present a comprehensive evaluation of the proposed cross-modal tracking method on the VT-Tiny-MOT dataset. We  first present details about the implementation of our method. Subsequently,  we conduct ablation study to analyze the effectiveness of different components of our proposed method. Finally, we compare our method with other state-of-the-art tracking methods.

\subsection{Implementation Details}

\subsubsection{Evaluation metrics}
To evaluate the performance of multi-object trackers in RGBT videos on the VT-Tiny-MOT benchmark, we employ several common quantitative metrics.  These metrics provide insights into various aspects of tracking performance. The metrics used for evaluation include Higher-Order Tracking Accuracy (HOTA) ,  Multi-Object Tracking Accuracy (MOTA), Multi-Object Tracking Precision (MOTP), ID-F1 Score (IDF1), ID switch (IDs), Mostly tracked targets (MT), Mostly lost targets (ML), False Positive (FP), and False Negative (FN). MOTA, MOTP, and IDF1 are employed to assess the overall performance of each tracker, while IDs, MT, ML, FP, and FN are reported to facilitate further comparison among the trackers.  The symbol $\uparrow$ after the metric means higher is better, while $\downarrow$ means lower is better. To ensure a more effective evaluation, considering the small size of the targets, we set the Intersection over Union (IOU) threshold to 0.3. 
\subsubsection{Experimental Settings} 
In the experiment, we utilized the PVTv2 as the embedding layer and employed the pretrained model to expedite the convergence of {HGT-Track} training. We computed the loss separately for each modality, and the sum of the losses was backpropagated to update the network parameters. The downscale ratio of the image feature was set to 4 to enable training on limited GPU computing resources. The detection threshold for both modalities was set to 0.4. All experiments were conducted on an RTX 3090 GPU. We set the graph layer to 1 and the spatial radius to 20. The training converged at around 20 epochs.

\subsection{Ablation Study}

\subsubsection{Heterogeneous Graph Transformer}

To evaluate the effectiveness of our Heterogeneous Graph Transformer (HGT) information integration module, we conducted experiments by training our model using varying numbers of HGT layers in the HGT encoder,  {denoted as `l-layer HGT', where `l' represents the number of HGT layers. The evaluation results for the visible and thermal test sequences are presented in Table \ref{res2}. Additionally, we trained a variant tracker without HGT, referred to as no-HGT.}

 {
Our HGT-based model consistently outperformed the variant `no-HGT' across most evaluation metrics. Specifically, compared to `no-HGT', `1-layer HGT' demonstrated an improvement of  $6.7/4.1 \%$ at MOTA  on  the visible/thermal test sequences, respectively.  Furthermore, the identification probability assessed by IDF1 exhibited enhancements of $12.4/10.2 \%$ in both modalities. With the number of HGT layer increases, the performance of `l-layer HGT' consistently surpasses `no-HGT'  , demonstrating the effectiveness of HGT.}
\subsubsection{ {Multi-Modal information integration}}
 {
 To demonstrate the effectiveness of multi-modal information integration, we introduced the variant  `l-layer HGT-s' that solely relies on single modality information  by excluding the edges $DH$ in `l-layer HGT'.}
 {
	As shown in Table \ref{res2}, the `3-layer HGT-s' variant exhibited an enhancement of $0.5/2.4 \%$ at MOTA on the visible and thermal test sequences, respectively, compared to `no-HGT'. This improvement demonstrates the effectiveness of incorporating temporal information.}
 {
	Furthermore, `l-layer HGT' showed superior performance to `l-layer HGT-s' across different numbers of HGT layers, which demonstrates the effectiveness of HGT in integrating information from visible and thermal modalities. 
}

\begin{figure*}[t]
	\centering
	\vspace{-.15in}
	\includegraphics[width=0.9\textwidth]{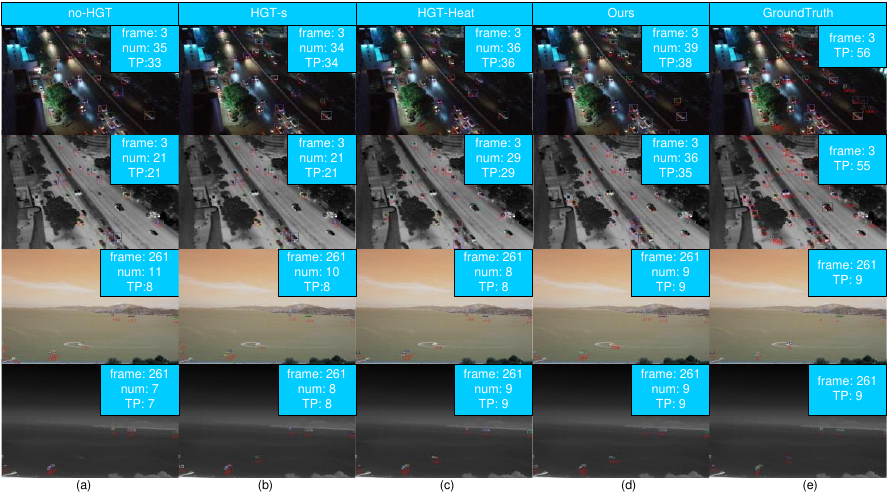}
	\caption{Qualitative results of our proposed method with different settings on two selected sequence pairs,  namely "Night Street" (upper two rows) and "Afternoon Sea" (bottom two rows),  from the test set of VT-Tiny-MOT.  The columns (a)-(d) display the results of different trackers. Our proposed tracker can generate more robust tracking results under such challenging scenarios.}

	\label{AQ}
\end{figure*}
\begin{table*}[tbp]
	
	\caption{Ablation Study on the effectiveness of HGT  {and multi-modal information integration. Quantitative results on the `Visible' and `Thermal' testing sequences of the dataset VT-Tiny-MOT are displayed,  with the top results highlighted in bold.  The column TI represent the usage of temporal information.} }
		\centering
			\tabcolsep=0.03\columnwidth
	\resizebox{\linewidth}{!}{
		\renewcommand\arraystretch{1}
		\begin{tabular}{c|lccccccccccc}
			\hline
			&\multicolumn{1}{l}{Method} & {Mod}& {TI}   &  {HOTA}$\uparrow$  & MOTA$\uparrow$  & MOTP$\uparrow$ & MT$\uparrow$ & ML $\downarrow$ &FP$\downarrow$&FN $\downarrow$ &IDF1$\uparrow$ &IDs$\downarrow$ \\
			\hline
			\multirow{9}*{\rotatebox{90}{Visible}}&no-HGT       & {V}& {\ding{56}} & {22.8}& 25.3      &  {54.2}     & 125   &  459     & 13669      & 117576    &30.8&2972 \\
			&1-layer HGT-s & {V } & {\checkmark } & {24.9} & 27.2      & 55.0      & 156  & 430      & 16860      &  110652   & 33.6&3375 \\
			&2-layer HGT-s & {V }& {\checkmark}   & {23.5} & 24.0      & 55.1      &140   &  429     &  20459     &112799   &31.9 & 3334 \\
			&3-layer HGT-s & {V} & {\checkmark }  & {24.7}  &  25.8     & 55.2      & 153   & 421   &  19025     & 111115  &34.6&3277 \\
			&4-layer HGT-s & {V} & {\checkmark}    & {20.1} &  14.7     & \textbf{59.9}      & 86   & 624   &  \textbf{6977}     & 144523  &21.5&1914 \\
			&1-layer HGT & {V+T}  & {\checkmark}  & {29.1} &  {32.0}   & {55.6}      & {226}      & {444}   & {11594}      &  {109500}     &  {43.2}     & 1223\\
			&2-layer HGT & {V+T}  & {\checkmark} & {\textbf{29.4}} & 31.6      &  55.2     &  \textbf{268}  & 406      & 16787      & \textbf{104812}   & 43.7&1288  \\
			&3-layer HGT & {V+T} & {\checkmark}   & {\textbf{29.4}} &  \textbf{32.1}     &55.1      & 257   &\textbf{399}       & 15595      & 105060   &\textbf{43.8}&1379 \\
			&4-layer HGT & {V+T} & {\checkmark}  & {28.9} &  29.5     & 55.9      & 211   &422       & 13835      & 111826   &41.2&\textbf{1053} \\
			\hline
			\multirow{9}*{\rotatebox{90}{Thermal}}&no-HGT  & {T} & {\ding{56}}    & {18.3}& 17.2      & 52.1    & 60   &  641     & {8395}      & 146650    &25.1&1252 \\
			&1-layer HGT-s & {T}  & {\checkmark}   & {18.3}& 14.5      & 50.9      & 66 &  618     & 13307      & 146580    &24.9 & 1348\\
			&2-layer HGT-s & {T} & {\checkmark }   & {18.3} & 17.4      & 50.2      &72   &618       & 10632      &143687   &25.9 & 1496 \\
			&3-layer HGT-s & {T}  & {\checkmark}   & {20.5}& 19.6      &  53.1     & 90   & 594      & 11162      &138768   &29.1& 1736\\
			&4-layer HGT-s & {T} & {\checkmark}   & {14.0} & 8.1      &  \textbf{53.8}     & 35   & 742      & \textbf{5330}      &167064   &13.8& 1048\\
			&1-layer HGT & {V+T} & {\checkmark}    & {23.1} &  {21.3}   & {51.5}      & {155}      & {588}   & 12516      &  {135197}     & \textbf{35.3}     & 844\\
			&2-layer HGT & {V+T} & {\checkmark}   & {21.3}&  16.6     &  {49.6}     &  87  & 606      &14913       &141765   &31.0&  \textbf{605}\\
			&3-layer HGT & {V+T} & {\checkmark} & {22.3} & 19.1      &  50.7     & 103   & 574      & 16928     &134726   &33.0& 997 \\
			&4-layer HGT & {V+T} & {\checkmark} & {\textbf{24.8}} & \textbf{22.1}      &  50.9     & \textbf{159}   & \textbf{556}      & 15695     &\textbf{130493}   &33.7& 725 \\ 
			\hline
			\end{tabular}}%
	\label{res2}%
\end{table*}%

\subsubsection{ReDet}
The ReDet module is proposed to promote consistency of object states between the two modalities.  We compare our tracker with the variant without ReDet  module (`Ours w/o ReDet') and another variant that utilizes heatmap-based re-detection, denoted as `HGT-Heat'.  `HGT-Heat'  performs re-detection of lost targets by employing the heatmap associated with the search region $SR$. 

The tracking performance is reported  in Table \ref{res3}. 
`Ours w/o ReDet' achieves the lowest false positive (FP) rate  due to a trade-off between FP and FN since neither of the re-detection methods can fully recognize the true target.  `HGT-Heat'  exhibits poor performance with with  $-0.05 \%$ in MOTA, due to a significant increase in false positives resulting from the lack of identity information in the heatmap. In contrast, `Ours w/ ReDet' effectively re-detects the lost targets using the correlation score map  followed by the verification of target. This results in an improvement of  $4.3 \%$ in MOTA and $9.8 \%$ in IDF1 as compared to`Ours w/o ReDet',  demonstrating the effectiveness of  ReDet module.

Furthermore, we illustrate the results obtained by `Ours w/ ReDet' under two distinct scenarios in column (d)  of Fig. \ref{AQ}. In the sequence pair ``Afternoon Sea", two ships are occluded by splashing water in the visible modality, while they are accurately recognized in the thermal modality. Our proposed tracker successfully redetect the lost target in visible modality through the ReDet module,  addressing the occlusion challenge.

\begin{table}[tbp]
		\tabcolsep=0.005\columnwidth
	 	\caption{Ablation studies on the effectiveness of ReDet module. Quantitative results on the `Visible' and `Thermal' testing sequences of the dataset VT-Tiny-MOT are displayed,  with the top results highlighted in bold.}
		\centering
	\resizebox{\columnwidth}{!}{
	\renewcommand\arraystretch{1.22}
\begin{tabular}{c|lccccccccc}
	\hline
	&\multicolumn{1}{l}{Method}    & HOTA$\uparrow$   & MOTA$\uparrow$  & MOTP$\uparrow$ & MT$\uparrow$ & ML $\downarrow$ &FP$\downarrow$&FN $\downarrow$ &IDF1$\uparrow$ &IDs$\downarrow$ \\
		\hline
		\multirow{3}*{\rotatebox{90}{Visible}}&Ours w/o ReDet      &24.8&27.7    & \textbf{55.7}      & 134      & 485      & \textbf{9082}      & 118231&33.4&2749  \\
		& HGT-Heat           &28.5 & -5.4      &{ 54.5   }   &  \textbf{267}     & \textbf{386}      & 69898 &\textbf{99332} &17.5& 20334 \\
		&Ours w/ ReDet            &\textbf{29.1} &\textbf{32.0}       &55.6       &  226     &   444    &11594 &{109500} & \textbf{43.2}& \textbf{1233} \\
		\hline
		\multirow{3}*{\rotatebox{90}{Thermal}}&	Ours w/o ReDet      &20.3&19.8   & \textbf{51.9}     & 88      & 614      & \textbf{9468}      & 140232&28.8&1492  \\
		& HGT-Heat           &15.3 & -10.7     &{ 50.5   }   &  \textbf{180}     & \textbf{497}      & 72355 &\textbf{121539} &18.5& 14945 \\
		&Ours w/ ReDet            &\textbf{23.1} &\textbf{21.3}       &51.5       &  155     &   588    &12516 &{135197} & \textbf{35.3}& \textbf{844} \\
		\hline
	\end{tabular}}%
	\label{res3}%
\end{table}%

\begin{table*}[tp]
	
	\tabcolsep=0.004\linewidth
	\renewcommand
	\arraystretch{1.05}
	 \caption{Quantitative results on the `Visible' and `Thermal' testing sequences of the dataset VT-Tiny-MOT  are displayed, with the top and second-top results highlighted in red and bold, respectively.  The GNN-based tracking methods marked with an asterisk (*) have been modified and retrained before evaluated on the VT-Tiny-MOT dataset.}
	\centering
	\resizebox{\linewidth}{!}{
		\begin{tabular}{|c|lccc|ccccccccccc|}
		   \hline
			&{Methods} &{FW}&Mod&Backbone & HOTA $\uparrow$& MOTA $\uparrow$ &MOTP $\uparrow$&MT $\uparrow$&ML $\downarrow$& FP  $\downarrow$   & FN  $\downarrow$   & IDF1 $\uparrow$ & IDs  $\downarrow$  &FPS $\uparrow$ &Params $\downarrow$   \\ \hline

			 \multirow{16}*{\rotatebox{90}{Visible}} &DeepSORT \cite{wojke2017simple}&TBD&V&ResNet50&23.4&12.4&68.8& 15& 623& 7043&148453&20.7&2062 &28.8&41.3  \\  
			
			  &Tracktor \cite{bergmann2019tracking} & TBD&V&ResNet50&23.3& 3.0& 64.2 & 152& 605& 28494& 144647&22.5&1283  &10.6& 67.2    \\
			
			 &ByteTrack \cite{zhang2022bytetrack} &TBD&V&ResNet50&\textbf{26.3} &  13.3&  68.7  &152& 625&6670  &148259& {26.3}& 844  &\textbf{38.4}& 41.3           \\
			
			  &OCSORT \cite{cao2022observation} &TBD&V&ResNet50&25.8 &  10.6& 68.0&{156}& {612} & 11940 & {146690} & 25.7& 2034  &24.3&41.3 \\
			
			&MPNTrack * \cite{braso2020learning} & TBD&V&ResNet50&23.0 & 3.5 & 62.4 &  168& 594&  29528 &  143452 & 24.6 & 418 &   -&21.0 \\
			  &DSFNet \cite{xiao2021dsfnet} &TBD&V&DLA34&16.3   &  10.4     & 64.3   & 32  &  747 & \textcolor{red}{911}  & 159954 & 16.6 & \textcolor{red}{290} &10.5&\textcolor{red}{17.0} \\
		
			  &CenterTrack \cite{zhou2020tracking}&JDT&V&DLA34&11.8 & 10.3   & \textcolor{red}{71.6}    &  16& 680    &\textbf{1859}     &  154977   &8.8  &  3933  &\textcolor{red}{41.0}&\textbf{19.7} \\
		
			  &FairMOT \cite{zhang2021fairmot} &JDT&V&DLA34&22.7 & 12.1   & {67.6} & 76 &688  &  2721     &  153421   &  22.4  & 1960 &21.7&20.1 \\
			
			  &TraDes \cite{wu2021track} &JDT&V&DLA34&20.0  & 8.8    &66.0  &  82     &  691& 9938  &  153532     & 20.4      &  428 &30.3&20.9 \\
			  &GSDT * \cite{wang2021joint}& JDT&V&DLA34&22.7 & 8.9 & \textbf{71.3} &  76& 688&  2722 &  153415 & 22.4 & 1950 & 1.3&21.0 \\
		
			  &Transcenter \cite{xu2022transcenter} & JDT&V&PVTv2&5.0  & -4.0 &  46.6 & 11& 856 & 10522 &  176028     & 2.1      &  \textbf{349} &22.3&35.0 \\
			\rowcolor{gray!10}\cellcolor{white}&{HGT-Track-V} (Ours)&JDT&V&PVTv2 &23.1 &  \textbf{27.2}   & {55.0}&{156}      & \textcolor{red}{430}   & 16860      &  \textbf{110652}     &  \textbf{33.6}     & 3375 &15.2&30.7 \\
			\cline{2-16}
			  &ProbEn-E+SORT \cite{chen2022multimodal} &TBD&V+T&ResNet50 &20.0  &9.5   & 66.7&138     & 542   & 19024      & 134952    &17.4& 8758 &-&60.3\\
			  &ProbEn-M+SORT \cite{chen2022multimodal}&TBD&V+T&ResNet50 &\textbf{26.3}  &  24.7   &67.4&\textbf{211}      &450   & 12108      &  111365     &  26.1     & 11930 &  -&120.6\\
			
			  &UA-CMDet+SORT \cite{sun2022drone}&TBD&V+T&ResNet50 &24.1  &8.6 & 65.1&190     & 464 & 36612      &121063& 23.2  & 6696 &  -&139.2\\
			\rowcolor{gray!10}\cellcolor{white}&{HGT-Track} (Ours)&JDT&V+T&PVTv2 &\textcolor{red}{29.1}  &  \textcolor{red}{32.0}   & {55.6}&\textcolor{red}{226}      & \textbf{444}   & 11594      &  \textcolor{red}{109500}     &  \textcolor{red}{43.2}     & 1223 &13.2&30.7 \\ \hline
		
			 \multirow{16}*{\rotatebox{90}{Thermal}}   &DeepSORT \cite{wojke2017simple} & TBD&T&ResNet50&21.9 &  8.4  & \textbf{73.7}& 142& 689  & 9859&  160768 & 17.2 & 2158 &28.7&41.3 \\
			 &Tracktor \cite{bergmann2019tracking} & TBD&T&ResNet50&24.5&5.8&69.5&153&661&22465&154425&2.7&881 &11.6&67.2 \\
			 &ByteTrack \cite{zhang2022bytetrack} & TBD&T&ResNet50&\textbf{25.5} &  9.2     &\textcolor{red}{73.9} & 150 & 692 & {9835}& 160458 &  22.6&{915} &\textbf{34.1}& 41.3 \\
			 &OCSORT \cite{cao2022observation} & TBD&T&ResNet50&24.4 & 0.9& {71.6} & 153 &  {655} &  27143 & {156803} & {21.4}  & 2921 &17.5&41.3 \\
				& MPNTrack * \cite{braso2020learning} & TBD&T&ResNet50&18.8 & -9.1 & 62.8 &  152& 635&  53635 &  151498 &19.0 & 847 &   -&21.0 \\
			 &DSFNet \cite{xiao2021dsfnet} &TBD &T&DLA34&20.5& 13.7 & 65.3 & 42      & 758      &  3678     & 158828      & 23.6      & \textbf{307} &10.2&\textcolor{red}{17.0} \\
			 &CenterTrack \cite{zhou2020tracking}  &JDT&T&DLA34&15.0 & \textbf{18.8}& {72.2} & 53 & 591 & 3089 & 143279 & 13.3 & 6194 &\textcolor{red}{39.9}&\textbf{19.7} \\
			& FairMOT \cite{zhang2021fairmot} &JDT&T&DLA34&15.3 & 8.0   & 62.2 & 45 &  785   & \textbf{1676}  & 170889 &14.9  & 928 &19.8&20.1 \\
			&TraDes \cite{wu2021track}&JDT&T&DLA34&\textcolor{red}{26.0}  & 15.7  & 67.0 &  153     & 609    &  15271     &  143153     & \textbf{30.5}      & 532 &28.3 &20.9  \\
			& GSDT * \cite{wang2021joint}& JDT&T&DLA34&15.3 & 8.9 & 62.3 &  45& 785&  {1678} &  170891 & 14.9 & 927 & 0.9&21.0 \\
			& Transcenter \cite{xu2022transcenter}& JDT&T&PVTv2&4.8 & 0.2 & 62.3 &  2& 909&  \textcolor{red}{1628} &  186443 & 1.6 & \textcolor{red}{129} &22.5&35.0 \\
			\rowcolor{gray!10}\cellcolor{white}&{HGT-Track-T} (Ours)&JDT&T&PVTv2 &18.3  &  {14.5}   & {50.9}&{66}      & {618}   & 13307      & {146580}     &  {24.9}     & 1348 &15.2&30.7 \\
		\cline{2-16}
			 &ProbEn-E+SORT \cite{chen2022multimodal} &TBD&V+T&ResNet50 &19.3  & 14.0   &69.8&126& 630   & 10229 &144052& 17.0 & 7981 &	  -&60.3 \\
			 &ProbEn-M+SORT \cite{chen2022multimodal} &TBD&V+T&ResNet50 &{20.7}  & {18.3}  &70.4&\textcolor{red}{250}     & \textcolor{red}{428}   & 9649      &  \textcolor{red}{110327}    & 16.3   &34042 &  -&120.6 \\
			 &UA-CMDet+SORT \cite{sun2022drone}&TBD&V+T&ResNet50 &19.8 & 10.5 &66.6&\textbf{247}&\textbf{466}   & 24452  &\textbf{120005} & 15.6  & 24390 &  -&139.2\\
			\rowcolor{gray!10}\cellcolor{white}&{HGT-Track} (Ours)&JDT &V+T&PVTv2 &{23.1} &\textcolor{red}{21.3}    & {51.5}      & {155}      &  {588}  & 12516      &  {135197}     &  \textcolor{red}{35.3}     & 844 &13.2&30.7 \\ \hline
			
		\end{tabular}}
	\label{res1}%
\end{table*}%
\subsubsection{The classification branch}\label{cb}
In the context of detection methods, the classification branch is commonly used to increase the inter-class difference between objects and decrease the intra-class difference. {HGT-Track} leverages the detection features as the ReID features of the target to associate targets across different time steps.

To explore the effectiveness of  classification branch within the remote sensing domain, we trained the detector using a binary classification network that was specifically designed to discriminate the target from background. This particular variant is referred to as  `Single'. The detailed tracking results are listed in Table \ref{res4}. A comparison with the tracking results obtained by the single-class model `Single' reveals that {HGT-Track} achieved an improvement of $2.4/2.6 \%$ in terms of MOTA for the visible and thermal modalities, respectively. These improvements effectively demonstrate the contribution of multi-classification branch to the overall performance of the proposed tracker.
\begin{table}[tbp]
		 
		\tabcolsep=0.005\columnwidth
		\caption{Ablation studies on the effectiveness of multi-calssification branch.  Top results are highlighted in bold. Quantitative results on the `Visible' and `Thermal' testing sequences of the dataset VT-Tiny-MOT are displayed,  with the top results highlighted in bold.}
		\centering
		\resizebox{\columnwidth}{!}{
		\renewcommand\arraystretch{1.5}
		\begin{tabular}{c|lccccccccc}
			\hline
			&\multicolumn{1}{l}{Method}     & HOTA$\uparrow$  & MOTA$\uparrow$  & MOTP$\uparrow$ & MT$\uparrow$ & ML $\downarrow$ &FP$\downarrow$&FN $\downarrow$ &IDF1$\uparrow$ &IDs$\downarrow$ \\
			\hline
			\multirow{2}*{\rotatebox{90}{Visible}}&Single  &26.4&29.6       & \textbf{56.5}      & 184      &\textbf{ 400 }     &  16133     &\textbf{105952} &36.1& 4460\\
			&Ours     &\textbf{29.1} &  \textbf{32.0}   & {55.6}      & \textbf{226}      & {444}   & \textbf{11594}      &  {109500}     &  \textbf{43.2}     & \textbf{1233}\\
			\hline
			\multirow{2}*{\rotatebox{90}{Thermal}}	&Single  &20.8 & 18.7      &  {49.6}     & 95      &  \textbf{558}     & 15783      &135569 &30.4&1945 \\
			&	Ours     &\textbf{23.1} &  \textbf{21.3}   & \textbf{51.5}      & \textbf{155}      & {588}   & \textbf{12516}      &  \textbf{135197}     &  \textbf{35.3}     & \textbf{844}\\
			
			\hline
			\end{tabular}}%
	\label{res4}%
\end{table}%
\subsection{Discussion}
\subsubsection{Detection  {and matching} threshold}

\begin{figure}[tbp]
	\centering
	\includegraphics[width=\linewidth]{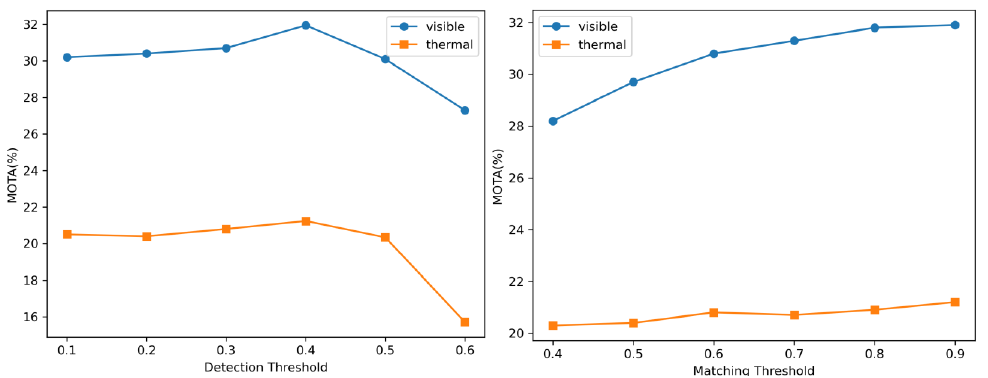}
		 \caption{Illustration of the influence of the detection threshold $\tilde{Det} $ and matching threshold  $\tau$ .}
	\label{dt}
\end{figure}

 {In the ReDet module, we propose two principles for checking the target position before  tracklet generation.} Lowering the detection threshold $\tilde{Det}$ allows for the detection of more targets with low confidence, but it also increases the number of false alarms.  {On the other hand, a higher $\tilde{Det}$ would decrease the number of true positives.}   The impact of the detection threshold on tracking performance is illustraed in Fig. \ref{dt}. While having more tracking nodes may facilitate better information aggregation from the other modality, it does not necessarily lead to improved tracking performance. Our sparse information fusion approach relies on accurate detection results, and increasing the number of false alarms degrades the overall tracking performance.

  {As for the IOU matching threshold $\tau$, a lower $\tau$ indicates a higher tolerance for differences in target detection between the two modalities. With an increase in $\tau$, there is a steady improvement in tracking performance in both modalities. This reflects that VT-Tiny-MOT is  aligned at the pixel level, and a higher IOU matching threshold will eliminate false alarms generated with the guidance of information from the other modality in the ReDet module.}

\subsubsection{Limitations}  In our proposed method, we track the target and maintain the tracklet by leveraging information from both modalities at the feature and decision levels. Through the designed HGT information fusion module, we are able to  effectively reduce false positives and false negatives. However, at a later stage, we prioritize maintaining the completeness of the tracklet, which may lead to a slight increase in false positives. This can be attributed to the heuristic setting of the ReDet module, which refines the target based on the physical prior that the tracklet should exhibit consistency in both modalities. However, false alarms are also retained under this setting.

\subsection{Comparison with Other Trackers}
In Table \ref{res1}, we present a comprehensive comparison of {HGT-Track} with other state-of-the-art trackers using the test set of VT-Tiny-MOT.  Our proposed tracker outperforms all currently dominant trackers in terms of overall performance, achieving the highest MOTA and IDF1 in both modalities.  For instance, compared to the ByteTrack tracker, {HGT-Track} shows an improvement of  $13/2.6 \%$ higher MOTA and $3.8/4.4 \%$ higher IDF1  for the visible and thermal modalities, respectively. Although our tracker does not achieve the best results on some specific metrics such as false positives (FP) and identity switches (IDs), it is important to consider the trade-off between false positives and false negatives (FN). Some methods such as FairMOT and Transcenter may achieve lower FP at the expense of high FN, resulting in the loss of many targets. This trade-off is a critical aspect for trackers.

 {
The tracking performance of three RGB-T trackers is also reported in Table \ref{res1}. {HGT-Track} demonstrates an improvement of $2.8 \%$ and $2.4 \%$ at HOTA compared to ProbEn-M+SORT in the two modalities, respectively. Trackers using RGB-T detectors can achieve lower false positives (FP) and false negatives (FN) compared to trackers using single-modality detector (Faster R-CNN). However, the tracking performance is limited due to the underutilization of temporal information, which is specifically addressed in our method.}

 {To demonstrate the effectiveness of HGT module in integrating spatial and temporal information, we trained our method  with single modality information, denoted as {HGT-Track-V and HGT-Track-T}.  Compared with the graph-based tracker GSDT, {HGT-Track-V} achieves higher tracking performace (MOTA: $18.3 \%\uparrow$) and higher tracking speed (FPS: $13.9\uparrow$). GSDT crops target patches from past frames to enhance current features. However, the crop operation costs additional time and may introduce more background noise considering the limited size of target. For the offline graph-based method MPNTrack, our method also achieved an improvement of $23.7/23.6 \%$ improvement at MOTA in visible and thermal modalities, respectively. The main reason is that the MPNTrack associate target depending on discriminative target ID feature, which is hard to learn in remote sensed scenarios.}

Our proposed method is specifically designed for multi-modal data and leverages the complementary information between modalities to improve tracking performance. The results demonstrate that our approach achieves strong tracking performance for various types of objects in different environmental conditions.  {Note that, to the best of our knowledge, existing public Multiple Object Tracking (MOT) datasets in the field of remote sensing are predominantly single-modal. Therefore, comparisons with other state-of-the-art methods on single-modal public datasets are not included to ensure fairness.}


\section{Conclusion}
In this paper, we introduce a unified large-scale visible-thermal benchmark named VT-Tiny-MOT for multiple tiny object tracking. We also  present a novel Joint Detection and Tracking (JDT) tracker called {HGT-Track}, which effectively integrates complementary information from both visible and thermal modalities to track objects.
{HGT-Track} utilizes  Heterogeneous Graph Transformer encoder and decoder to integrate information across modalities.  The information flows through our spatial confined heterogeneous graph, where the edge is built dependant on modality and spatial distance. This sparse information integration approach improves the efficiency of our tracking method and excludes pixel matching error between two modalities.  Additionally, we introduce the ReDet module to ensure the consistency of target trajectories between different modalities, effectively reducing false negatives.  Our proposed method achieves the best performance compared to other state-of-the-art trackers on the VT-Tiny-MOT dataset.


%
%
%
%


%

\bibliographystyle{IEEEtran}
\bibliography{scholar}

\vspace{-33pt}
\begin{IEEEbiography}[{\includegraphics[width=1in,height=1.25in,clip,keepaspectratio]{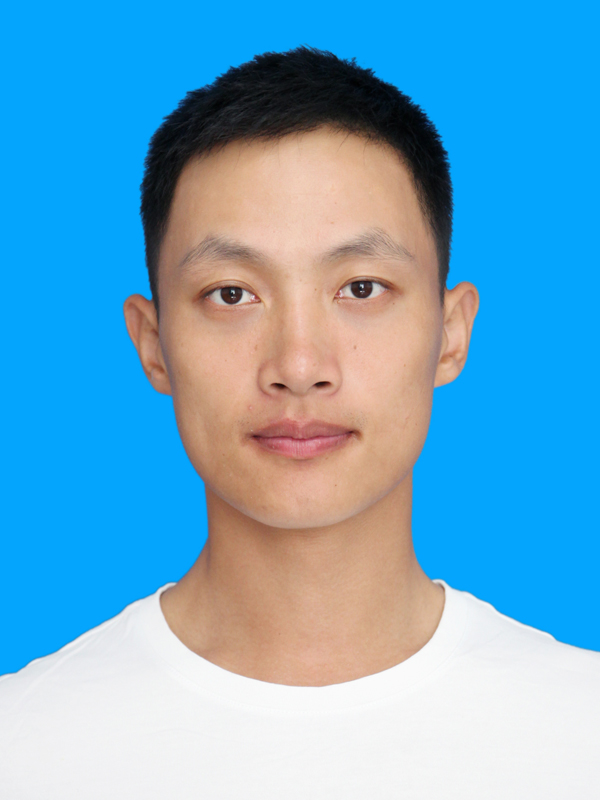}}]{Qingyu Xu}
	received the B.E. degree in Electrical Engineering from Shandong University (SDU), Jinan, China, in 2015, and the M.E. degree in Information and Communication Engineering from National University of Defense Technology (NUDT), Changsha, China, in 2021. He is currently persuing the Ph.D. degree with the College of Electronic Science and Technology, NUDT.  His current research interests include information fusion and object tracking.
\end{IEEEbiography}

\begin{IEEEbiography}[{\includegraphics[width=1in,height=1.25in,clip,keepaspectratio]{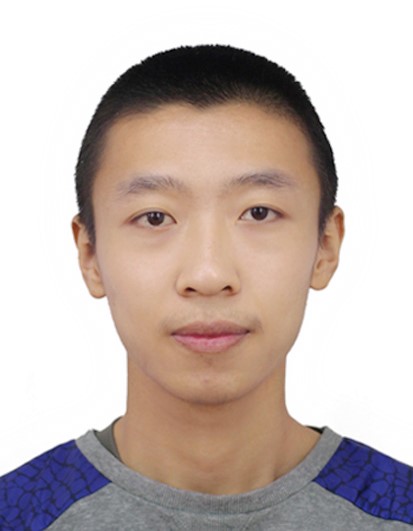}}]{Longguang Wang}
	received the B.E. degree in Electrical Engineering from Shandong University (SDU), Jinan, China, in 2015, and the Ph.D. degree in Information and Communication Engineering from National University of Defense Technology (NUDT), Changsha, China, in 2022. His current research interests include low-level vision and 3D vision.
\end{IEEEbiography}
\begin{IEEEbiography}[{\includegraphics[width=1in,height=1.25in,clip,keepaspectratio]{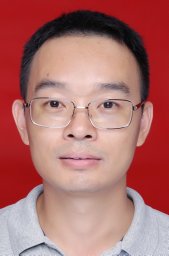}}]{Weidong Sheng}
	received the Ph.D. degree from the National University of Defense Technology (NUDT), Changsha, China, in 2011. He is currently an Associate Professor with the College of Electronic Science and Technology, NUDT. His current research interests include image processing and data fusion.
\end{IEEEbiography}
\begin{IEEEbiography}[{\includegraphics[width=1in,height=1.25in,clip,keepaspectratio]{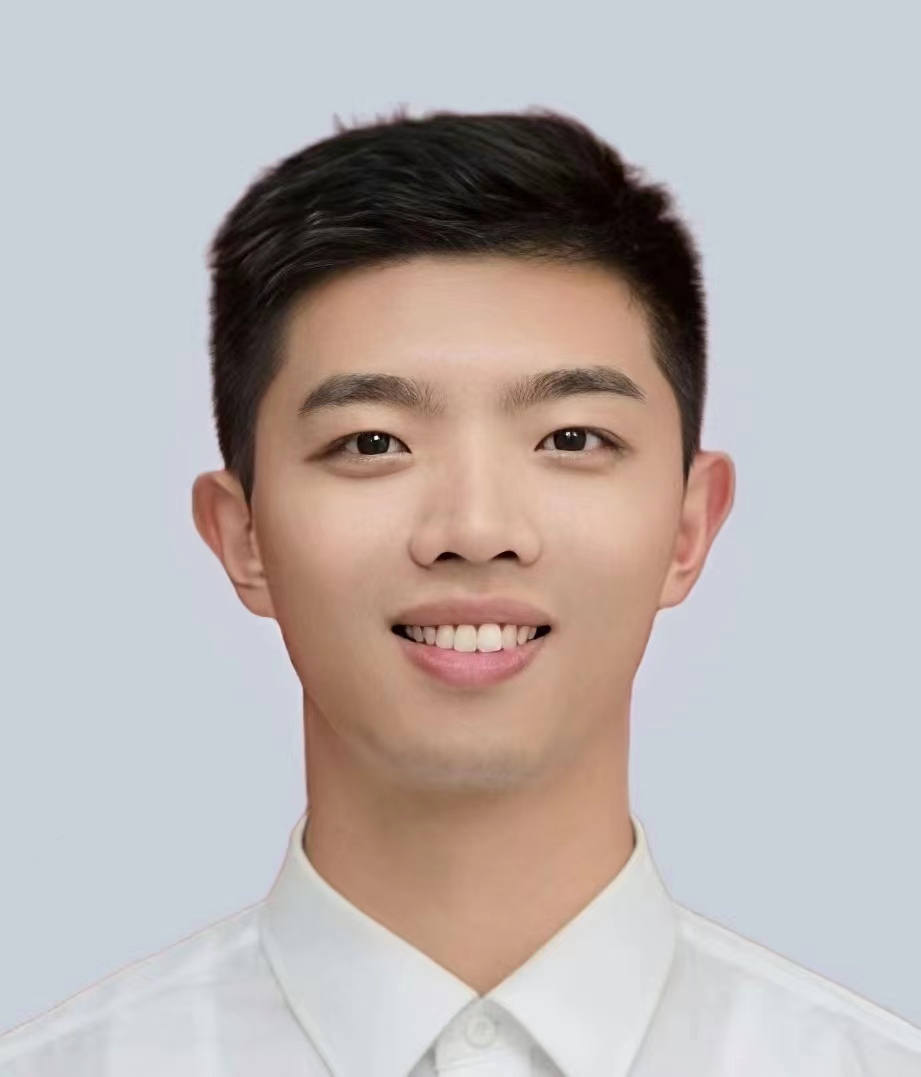}}]{Yingqian Wang}
	received the B.E. degree in
	electrical engineering from Shandong University,
	Jinan, China, in 2016, and the master’s and
	Ph.D. degrees in information and communication
	engineering from the National University of Defense
	Technology (NUDT), Changsha, China, in 2018 and
	2023, respectively.
	
	He is currently an Assistant Professor with the
	College of Electronic Science and Technology,
	NUDT. His research interests focus on optical
	imaging and detection, particularly on light field
	imaging, image super-resolution, and infrared small target detection.
\end{IEEEbiography}
\begin{IEEEbiography}[{\includegraphics[width=1in,height=1.25in,clip,keepaspectratio]{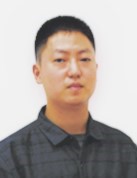}}]{Chao Xiao}
	received the BE degree in the communication
	engineering and the Ph.D. degree in information
	and communication engineering from the National
	University of Defense Technology (NUDT), Changsha,
	China in 2016 and 2023, respectively.  His research interests include deep learning,
	small object detection and multiple object tracking.
\end{IEEEbiography}
\begin{IEEEbiography}[{\includegraphics[width=1in,height=1.25in,clip,keepaspectratio]{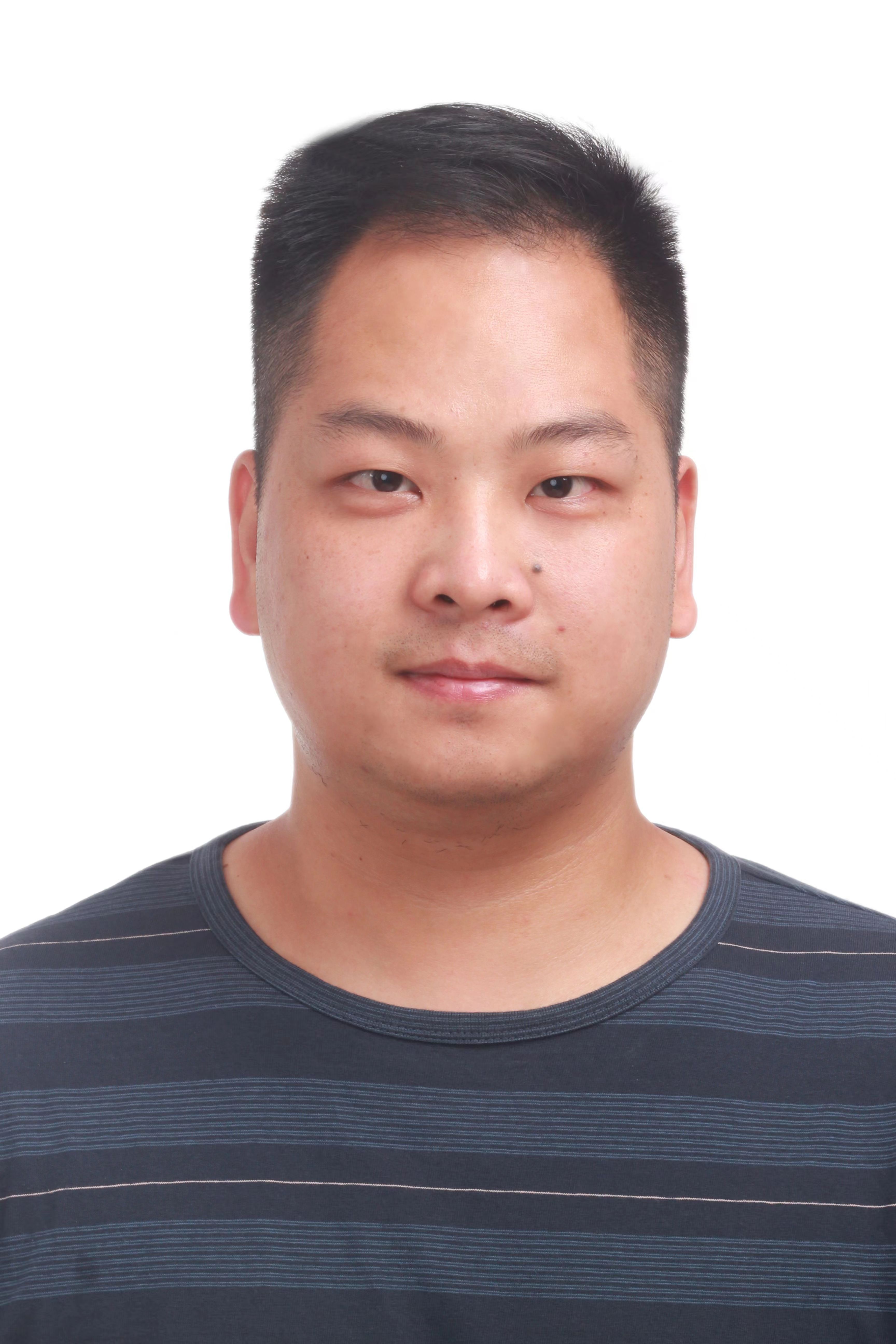}}]{Chao Ma}
	 received the Ph.D. degree in information and communication engineering from the National University of Defense Technology (NUDT), Changsha, China, in 2019. He is currently an Associate Researcher  with the College of Electronic Science and Technology, NUDT. His research interests include remote sensing processing, target tracking, information fusion and 3D vision.
\end{IEEEbiography}
\begin{IEEEbiography}[{\includegraphics[width=1in,height=1.25in,clip,keepaspectratio]{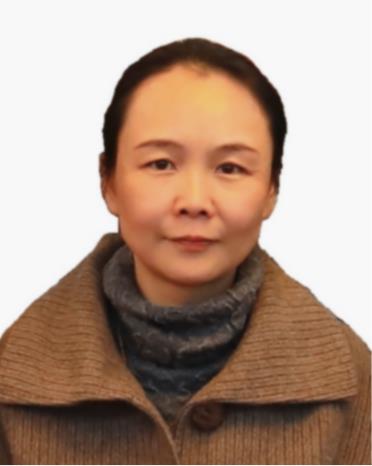}}]{Wei An}
	received the Ph.D. degree from the National University of Defense Technology (NUDT), Changsha, China, in 1999. She was a Senior Visiting Scholar with the University of Southampton,
	Southampton, U.K., in 2016. She is currently a Professor with the College of Electronic Science and Technology, NUDT. She has authored or co-authored over 100 journal and conference publications. Her current research interests include signal processing and image processing.
\end{IEEEbiography}

\vfill

\end{document}